\documentclass[letterpaper]{article} 
\usepackage{aaai25}  
\usepackage{times}  
\usepackage{helvet}  
\usepackage{courier}  
\usepackage[hyphens]{url}  
\usepackage{graphicx} 
\usepackage{natbib}  
\usepackage{caption} 
\usepackage{algorithm}
\usepackage{algorithmic}
\usepackage[T1]{fontenc}
\usepackage{amsmath}
\usepackage{amsfonts}
\usepackage{booktabs}
\usepackage{multirow}
\usepackage{makecell}
\usepackage{array}

\usepackage{newfloat}
\usepackage{listings}
\DeclareCaptionStyle{ruled}{labelfont=normalfont,labelsep=colon,strut=off} 
\floatstyle{ruled}
\newfloat{listing}{tb}{lst}{}
\floatname{listing}{Listing}
\title{ResAdapter: Domain Consistent Resolution Adapter for Diffusion Models}
\author{
    Jiaxiang Cheng,
    Pan Xie\thanks{Corresponding author},
    Xin Xia,
    Jiashi Li,
    Jie Wu, \\
    Yuxi Ren,
    Huixia Li,
    Xuefeng Xiao,
    Min Zheng,
    Lean Fu
}
\affiliations{
    ByteDance Inc., Beijing, China\\

    jiaxiangcc@gmail.com, xiepan.01@bytedance.com \\
    \{xiaxin.97,lijiashi,wujie.10,renyuxi.20190622,lihuixia,xiaoxuefeng.ailab\}@bytedance.com
}

\usepackage{bibentry}

\begin{document}

\maketitle

\begin{abstract}

Recent advancement in text-to-image models and corresponding personalized technologies enables individuals to generate high-quality and imaginative images.
However, they often suffer from limitations when generating images with resolutions outside of their trained domain.
To overcome this limitation, we present the resolution adapter \textbf{(ResAdapter)}, a domain-consistent adapter designed for diffusion models to generate images with unrestricted resolutions and aspect ratios.
Unlike other multi-resolution generation methods that process images of static resolution with complex post-process operations, ResAdapter directly generates images with the dynamical resolution.  
Especially, after learning a deep understanding of pure resolution priors, ResAdapter trained on the general dataset, generates resolution-free images with personalized diffusion models while preserving their original style domain.
Comprehensive experiments demonstrate that ResAdapter with only 0.5M can process images with flexible resolutions for arbitrary diffusion models.
More extended experiments demonstrate that ResAdapter is compatible with other modules for image generation across a broad range of resolutions, and can be integrated into other multi-resolution model for efficiently generating higher-resolution images.

\end{abstract}

\vspace{-0.5em}

\begin{links}
    \link{Code}{https://github.com/bytedance/res-adapter}
\end{links}

\vspace{-0.5em}

\begin{figure*}[tb]

  \centering
  \includegraphics[width=0.95\linewidth]{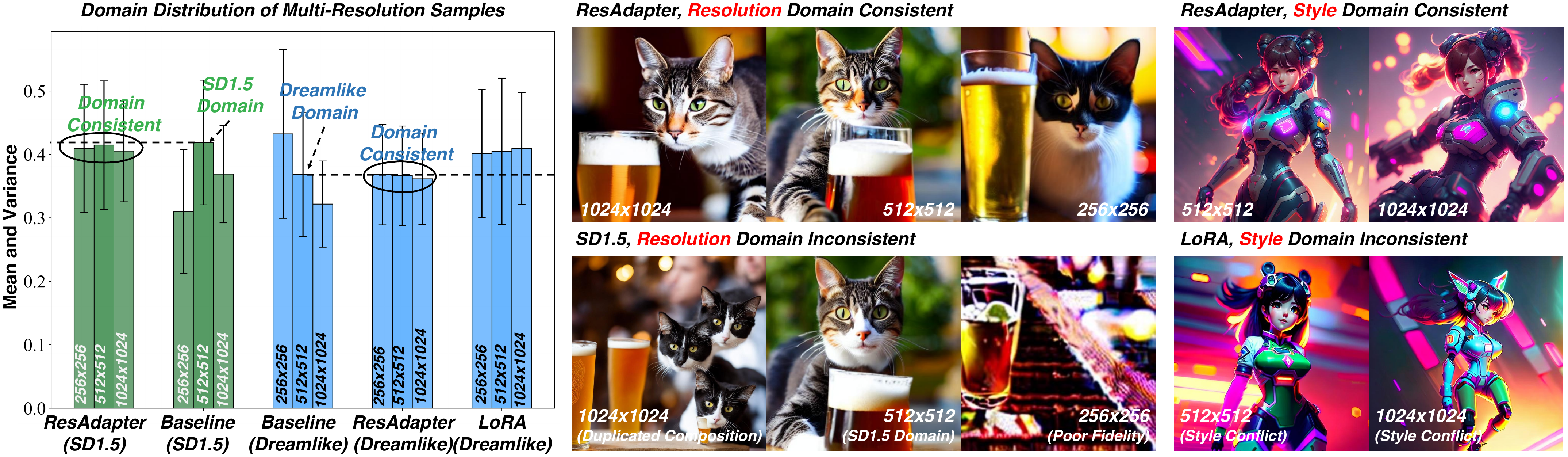}
  \caption{
  \textbf{Motivation}.
  We explore the domain distribution of images generated by SD1.5 and Dreamlike at resolutions of $256 \times 256$, $512 \times 512$ and $1024 \times 1024$. Dreamlike is the personalized diffusion model based on SD1.5.
  We find that \textbf{baselines} transform domains at resolutions of $256 \times 256$ and $1024 \times 1024$.
  The above ResAdapter and LoRA are both trained on the general dataset LAION-5B.
  \textbf{ResAdapter} keep domain consistent at different resolutions.
  But \textbf{LoRA} injects style priors from LAION-5B and influences the Dreamlike domain, resulting to low-quality images with the style conflict.
  }
  \label{fig:motivation}
  \vspace{-1.5em}

\end{figure*}

\section{1\quad Introduction}
\label{sec:intro}

Diffusion models~\cite{ho2020denoising,song2020score,song2020denoise} have experienced a remarkable surge in their capabilities and applications~\cite{lugmary2022repaint,meng2021sdedit,ramesh2022hierarchical}.
Among them, Stable Diffusion (SD)~\cite{rombach2022high} and SDXL~\cite{podell2023sdxl} are pre-trained models on the large-scale dataset LAION-5B~\cite{schuhmann2022laion}, having emerged as powerful generative models.
Additionally, the open-source community has been enriched by numerous personalized diffusion models from CivitAI~\cite{Civitai}, trained with DreamBooth~\cite{ruiz2023dreambooth} or Low-rank Adaptation (LoRA)~\cite{hu2021lora}.
They are capable of generating imaginative high-quality images at the training resolution (e.g., $512 \times 512$ for SD-based models and $1024 \times 1024$ for SDXL-based models) using the given prompts. 
However, they often suffer from limitations when generating images with resolutions outside of their trained domain. 
As shown in Figure~\ref{fig:motivation}, the SD-based model and the personalized diffusion model generate lower-resolution images (e.g., $256 \times 256$) with the poor fidelity and higher-resolution images (e.g., $1024 \times 1024$) with the poor framing and composition.
As a result, we can name this phenomena as the \textit{resolution domain inconsistent}. 

\begin{table}[tb]
\centering
\resizebox{0.95\columnwidth}{!}{%
\begin{tabular}{@{}l|c|c|c|c|c@{}}
\toprule
\bf{Method} & \bf{Type} & \bf{\makecell{Domain\\ Consistent}} & \bf{\makecell{Module\\ Compatible}} & \bf{\makecell{Training\\ Inexpensive}} & \bf{\makecell{Inference\\ Efficient}} \\ \midrule
ASD         & Train          & $\times$ & $\times$     & $\times$     & $\times$ \\
SDXL        & Train          & $\times$ & $\times$     & $\times$     & $\checkmark$ \\
Diffit      & Train          & $\times$ & $\times$     & $\checkmark$ & $\checkmark$ \\
MultiDiff   & Train-free          & $\times$ & $\times$     & $\checkmark$ & $\times$ \\
ElasticDiff & Train-free          & $\times$ & $\times$     & $\checkmark$ & $\times$ \\
MixtureDiff & Train-free          & $\times$ & $\times$     & $\checkmark$ & $\times$ \\
HiDiffusion & Train-free          & $\times$ & $\times$     & $\checkmark$ & $\checkmark$ \\
LoRA        & Few-shot          & $\times$ & $\checkmark$ & $\checkmark$ & $\checkmark$  \\ \midrule
ResAdapter  & Zero-shot          & $\checkmark$ & $\checkmark$ & $\checkmark$ & $\checkmark$ \\ \bottomrule
\end{tabular}%
}
\caption{\textbf{Comparison for ResAdapter and other methods}.
\underline{Domain Consistent}: resolution and style domain of generation image maintain consistent for arbitrary diffusion models.
\underline{Module Compatible}: compatible with other modules except diffusion models.
\underline{Training Inexpensive}: low-cost training.
\underline{Inference Efficient}: process images without repeated denoising steps and complex post-process operations.
}
\label{tab:compare}
\vspace{-1.5em}
\end{table}

Existing work is categorized into two main research directions to address this limitation. 
The first research line is train-free direction~\cite{jim2023mixture,bar2023multidiffusion,haji2024elastic,zhang2024hi}, represented by MultiDiffusion and ElasticDiffusion, where images with resolutions in their trained domain are repeatedly processed and then stitched together to generate images with flexible resolutions through overlap. However, these approaches often take longer inference time with complex post-process operations.
The second research line is straightforward. Fine-tuning models~\cite{zheng2024any} or additional parameters~\cite{hu2021lora} on a broader range of resolutions to empower diffusion models to generate resolution-free images. However, most personalized models in CivitAI~\cite{Civitai} do not provide details about their training datasets. Fine-tuning on the general dataset like LAION-5B~\cite{schuhmann2022laion} inevitably influences their original style domain, which is shown in Figure~\ref{fig:motivation}. We name this phenomena as the \textit{style domain inconsistent}.

\textbf{Can we train a plug-and-play resolution adapter to generate images with unrestricted resolutions and aspect ratio for arbitrary diffusion models?}
To answer this question, we decompose it into three dimensions.
(1) \textit{Resolution interpolation}: generate images with resolutions below the trained resolution of diffusion models.
(2) \textit{Resolution extrapolation}: process images with resolutions above the trained resolution of diffusion models.
(3) \textit{Style consistency}: generate images without transforming the original style domain of the personalized diffusion model.

We analyze the structure of diffusion model blocks~\cite{ronne2015unet}, finding that the attention and feed-forward are both content-sensitive layers, which are sensitive to the style information of images compared to resolution.
However, the convolution layers with fixed receptive file are resolution-sensitive, meaning they are easily influenced by the resolution of generation images.
Leveraging these finds, we present the resolution convolution LoRA (\textit{ResCLoRA}) for dynamically matching the receptive filed of convolution and the feature map size of images with flexible resolutions.
However, we find that as the resolution increases, the gap about the quality of generation images increases between LoRA and full fine-tuing. 
We attribute it into the inability of normalization in diffusion model blocks to adapt the statistical distribution of images in resolution extrapolation. 
According to this, we present the resolution extrapolation normalization (\textit{ResENorm}) for reducing the gap between LoRA and full fine-tuning in resolution extrapolation.
To enable the style domain consistency, we optimize the position of ResCLoRA and ResENorm insertions on diffusion model blocks to guide them to learn resolution priors ignoring the style information from the general datasets.

Through integrating these two optimized methods, we can train a plug-and-play domain-consistent resolution adapter (\textbf{ResAdapter}), which expands the range of resolution domain from diffusion models without transforming their original style domains.
Our main experiments demonstrate that after learning resolution priors, ResAdapter with only 0.5M can expand the generation resolution of SD-based personalized models from $128 \times 128$ to $1024 \times 1024$ and scale the resolution of SDXL-based personalized models from $256 \times 256$ to $1536 \times 1536$.
Our extensive experiments demonstrate that ResAdapter is compatible with other modules (e.g., ControlNet~\cite{zhang2023adding} for conditional generation, IP-Adapter~\cite{ye2023ip} for image generation based on the image prompt and LCM-LoRA~\cite{luo2023lcmlora} for accelerating generation), and even can be integrated into other multi-resolution models (e.g., ElasticDiffusion~\cite{haji2024elastic}) for efficiently generating $2048 \times 2048$ high-resolution images. Detailed comparison with other related work is summarized in Table~\ref{tab:compare}.
Our contributions can be summarized as follows:
\begin{itemize}
    \item We present a plug-and-play domain-consistent ResAdapter for generating images of resolution interpolation and extrapolation with diffusion models. 
    \item ResAdapter enables diffusion models of arbitrary style domain to generate images of unrestricted resolution and aspect ratio without transforming their style domain.
    \item ResAdapter is lightweight and without complex post-process operations. We can train it once for only 0.5M with low-cost consumption and efficiently inference resolution-free images.
    \item ResAdapter is compatible with other modules to generate images with flexible resolution, such as ControlNet, IP-Adapter and LCM-LoRA, even can optimize generation efficiency of other multi-resolution models.
\end{itemize}

\begin{figure*}[tb]

  \centering
  \includegraphics[width=0.95\linewidth]{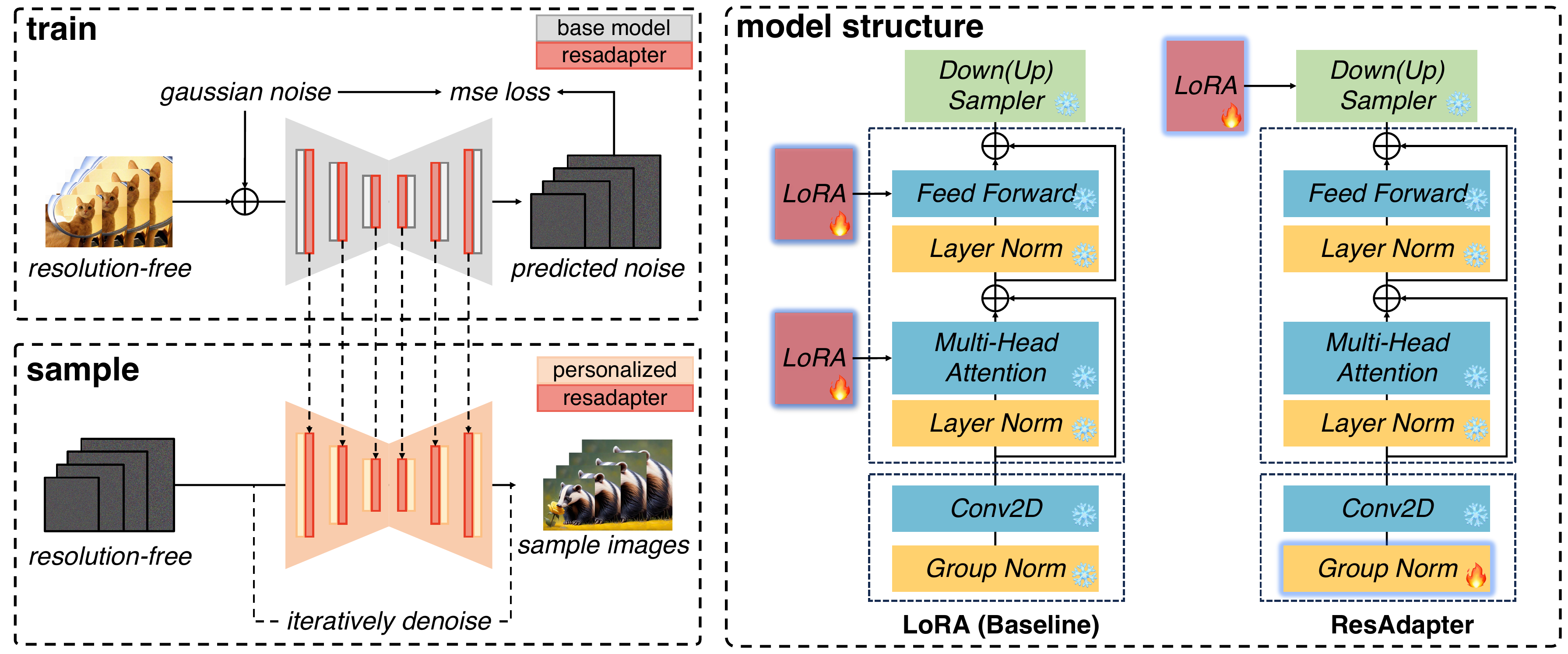}
  \caption{
  \textbf{Overview of ResAdapter}. 
  \textbf{Left}: Pipeline of ResAdapter.
  ResAdapter based on the frozen base model (e.g., SD or SDXL) learns resolution priors from mixed-resolution general datasets, and can be integrated into arbitrary personalized models to generate multi-resolution images.
  \textbf{Right}: Architecture comparison between ResAdapter and LoRA. Compared to LoRA, ResAdapter is only inserted to downsampler and upsampler, and it unfreezes the group normalization of resnet blocks.
  }
  \label{fig:framework}
  \vspace{-1.5em}
  
\end{figure*}

\section{2\quad Related Work}
\label{sec:related}

\subsection{2.1\quad Text-based Image Generation}
The rapid development of artificial intelligence generative component has attracted growing interest in text-to-image generation. 
GAN-based methods~\cite{alec2016dcgan,goodfellow2020generative,odena2017conditional} employ small-scale data for training but encounter challenges in adapting to large-scale data due to the instability of the adversarial training process. 
Autoregressive-based methods~\cite{van2017neural,esser2021taming,lee2022autoregressive} learn the latent distribution of discrete latent spaces but take more inference costs. 
Recently, diffusion models~\cite{song2020score,ho2020denoising,karras2022elucidating,dhariwal2021diffusion} has emerged as the state-of-the-art model in the text-to-image generation field.
The diffusion models represented by Stable Diffusion~\cite{rombach2022high} and SDXL~\cite{podell2023sdxl} contribute to high-resolution image generation. With the advent of personalized techniques~\cite{ruiz2023dreambooth,hu2021lora}, they are capable of generating imaginative images.
However, they still encounter limitations in resolution-free image generation.

\subsection{2.2\quad Resolution-free Image Generation}
Existing work of resolution-free image generation mainly utilizes post-processing to generate images beyond the training resolution. 
Mixture-of-Diffusers~\cite{jim2023mixture} and MultiDiffusion~\cite{bar2023multidiffusion} utilize pre-trained Stable Diffusion~\cite{rombach2022high} to generate $512 \times 512$ images multiple times, and overlap to generate high-resolution landscape images. But this also lead to duplicated objects.
ASD~\cite{zheng2024any} fine-tunes on multi-aspect ratio images, and generates high-resolution images throught implicit overlap. 
ElasticDiffusion~\cite{haji2024elastic} optimizes the post-processing process and can generate lower resolution images. 
Compared to these work, our ResAdapter does not require post-processing that take more inference time and can be integrated into any personalized model. ResAdapter can even be combined with these work to optimize inference time of generating higher resolution images.
\section{3\quad The Method}
\label{sec:method}

In this section, we delve into our proposed plug-and-play domain-consistent ResAdapter, which enables diffusion models of arbitrary style domain to generate images with unrestricted resolutions and aspect ratio.
First, we introduce ResCLoRA, which enables diffusion models to generate images with resolution interpolation.
Then, we introduce ResENorm, which compensates for the lack of capability about ResCLoRA in resolution extrapolation.
Finally, we present a simple multi-resolution training strategy, which can effectively make diffusion models generate images with flexible resolutions through only one ResAdapter.

\subsection{3.1\quad Resolution Interpolation}
\label{reslora}

LoRA~\cite{hu2021lora} enables the base model (e.g., SD~\cite{rombach2022high} and SDXL\cite{podell2023sdxl}) to generate high-quality style images. 
As shown in Figure~\ref{fig:framework}, LoRA is inserted into the query, value, key and output layers of the attention block to learn the style domain distribution of images. It is defined as $\mathcal{W}_{a}^{\prime} = \mathcal{W}_{a} + \Delta\mathcal{W} = \mathcal{W}_{a} + AB^T$, where $A \in \mathbb{R}^{m\times r}$, $B \in \mathbb{R}^{n \times r}$ are two rank-decomposition matrices, $r$ represents the rank of matrices.
However, LoRA trained on the general datasets can not be integrated into other personalized diffusion models, which influences their original style domain.
As shown in Figure~\ref{fig:motivation}, LoRA trained on LAION-5B~\cite{schuhmann2022laion} transforms the domain of the personalized model to the domain of SD1.5 and generates bad quality images with style conflicts.

ResCLoRA can be integrated into any personalized model to enable resolution interpolation for high-quality images without transforming the style domain. 
The reason that leads to the poor fidelity of images with resolution interpolation is that the convolution with the fixed receptive field is sensitive to the resolution of images.
According to this, ResCLoRA is inserted into the convolution layers of diffusion blocks to learn resolution priors. 
To prevent as much as possible ResCLoRA from capturing the style domain of the general datasets, it is only inserted into the convolution layers in downsampler and upsampler blocks.
We define ResCLoRA as $\mathcal{W}_{d}^{\prime}=\mathcal{W}_{d}+AB^T$  and $\mathcal{W}_{u}^{\prime}=\mathcal{W}_{u}+AB^T$, which is shown in Figure~\ref{fig:framework}.
Compared to the style information, the resolution information is low-level knowledge. 
Thus, ResCLoRA with only 0.4M can provide rich resolution priors for personalized models, adaptively adjusting the receptive field of convolution in diffusion blocks to match the feature map size of generation images while preserving their style domain.

\subsection{3.2\quad Resolution Extrapolation}
\label{resnorm}

Our initial experiment finds that there is a large gap as the resolution increases between LoRA and full fine-tuning. 
This means that only ResCLoRA does not enable the resolution extrapolation ability of the personalized model.
For example, ResCLoRA integrated into the diffusion model still generates higher-resolution images with poor framing and composition.
Inspired by LongLoRA~\cite{chen2024long}, we find that the failure of the resolution extrapolation is limited by the ability of normalization layers. 
Existing normalization layers can not adapt to the statistical distribution of feature maps of higher-resolution images.

However, we find that all normalization layers of diffusion models blcoks trained on LAION-5B~\cite{schuhmann2021laion} are not compatible with the other parameters of the personalized model, which still leads to low-quality images with poor style color.
In order to keep the original style domain of generation images, we need to maintain partial normalization layers of the personalized model.
As shown in Figure~\ref{fig:framework}, we only open group normalization of resnet layer, which is named as ResENorm. It not only reduce the gap about the resolution prior between ResCLoRA and full fine-tuning, but also helps retain the style domain of the personalized model. 
Additionally, we only train ResENorm in resolution extrapolation for better adapting to the statistical distribution of the feature map of higher-resolution images. 
After that, ResENorm with ResCLoRA can improve the poor framing and composition of generation images.
Especially, ResENorm only occupies 0.1M parameters but make effects for reducing the gap in resolution extrapolation.

\subsection{3.3\quad Resolution-Free Consistency Training}
\label{train}

To enable resolution-free domain consistent image generation for single ResAdapter, we propose a simple mixed-resolution training strategy based on our specific adapter design, as shown in Figure~\ref{fig:framework}. 
For SD~\cite{ramesh2022hierarchical}, we train on the mixed datasets with common resolutions from $128 \times 128$ to $1024 \times 1024$ with unrestricted aspect ratio. 
For SDXL~\cite{podell2023sdxl}, we train on the mixed datasets with common resolutions from $256 \times 256$ to $1536 \times 1536$ with unrestricted aspect ratio. 
In the training process, the base model is frozen, and only the ResAdapter is trainable.
The multi-resolution training strategy, along with the specific adapter design, enables ResAdapter to learn multi-resolution knowledge simultaneously while preventing catastrophic forgetting~\cite{masip2023continual,smith2023continual,gao2023ddgr} from full fine-tuning.

Our experiments find that the lower and higher resolution images (e.g., $128 \times 128$ and $1024 \times 1024$ for SD) are more difficult to train. 
To alleviate this phenomenon, we use a simple probability function to sample images at the different training resolution.
It is defined as $p(x)=|x-s|^2/\sum_{i}^{N} |x_i-s|^2$, where $s$ represents the standard training resolution (e.g., $512 \times 512$ for SD). 
This can improve the probability of selecting lower or higher resolution images resolution during the multi-resolution training process.
\section{4\quad Experiments}

In this section, we introduce the experimental setup and results. 
First, we describe the experimental setup in detail, including training details, evaluation metrics, and the selection of personalized models. 
And we show the main experimental results. We compare ResAdapter with other multi-resolution image generation models as well as the original personalized model.
Then we show the extended experimental results. That is the application of ResAdapter in combination with other modules. 
Finally, we perform ablation experiments about ResAdapter modules and alpha.

\subsection{4.1\quad Experimental Setup}
\label{expsetup}

\textbf{Training Details.}
We train ResAdapter using the large-scale dataset LAION-5B~\cite{schuhmann2022laion}.
Considering most structures of personalized models in the open-source community, we choose SD1.5~\cite{rombach2022high} and SDXL1.0~\cite{podell2023sdxl} as the base models. 
For SD1.5, we train on images with $128 \times 128, 256 \times 256, 384 \times 384, 768 \times 768$ and $1024 \times 1024$ resolutions. 
For SDXL, we expanded the training resolution range to $256 \times 256, 384 \times 384, 512 \times 512, 768 \times 768, 1280 \times 1280, 1408 \times 1408$, and $1536 \times 1536$.
Meanwhile, the training dataset contains images with different ratios such as 4:3, 3:4, 3:2, 2:3, 16:9 and 9:16.
For SD1.5 and SDXL, we both use a batch size of 32 and a learning rate of 1e-4 for training. 
We use the AdamW optimizer~\cite{kingma2014adam} with $\beta_1=0.95, \beta_2=0.99$. The total number of training steps is 20,000. 
Since ResAdapter is only 0.5M of trainable parameters, we train it for less than an hours on 8 $\times$ A100 GPUs.

\noindent \textbf{Evaluation Metrics.}
For experiments comparing  ResAdapter with the personalized model, we hire 5 humans to participating in the qualitative evaluation. 
For experiments comparing ResAdapter and the other multi-resolution image generation models, we refer to~\cite{haji2024elastic} and use Fréchet Inception Distance (FID)~\cite{heusel2017gans} and CLIP Score~\cite{hessel2021clipscore} as evaluation metrics. 
They evaluate the quality of the generated images and the degree of alignment between the generated images and prompts. 
For other multi-resolution generation models, we chose MultiDiffusion (MD)~\cite{bar2023multidiffusion} and ElasticDiffusion (ED) as baselines.

\noindent \textbf{Personalized Models.}
In order to demonstrate the effectiveness of our ResAdapter, we choose multiple personalized models from Civitai~\cite{Civitai}, which cover a wide domains range from animation to realistic photography. For personalized model information, see details in Appendix B.

\begin{figure*}[tb]
  \centering
  \includegraphics[width=0.95\linewidth]{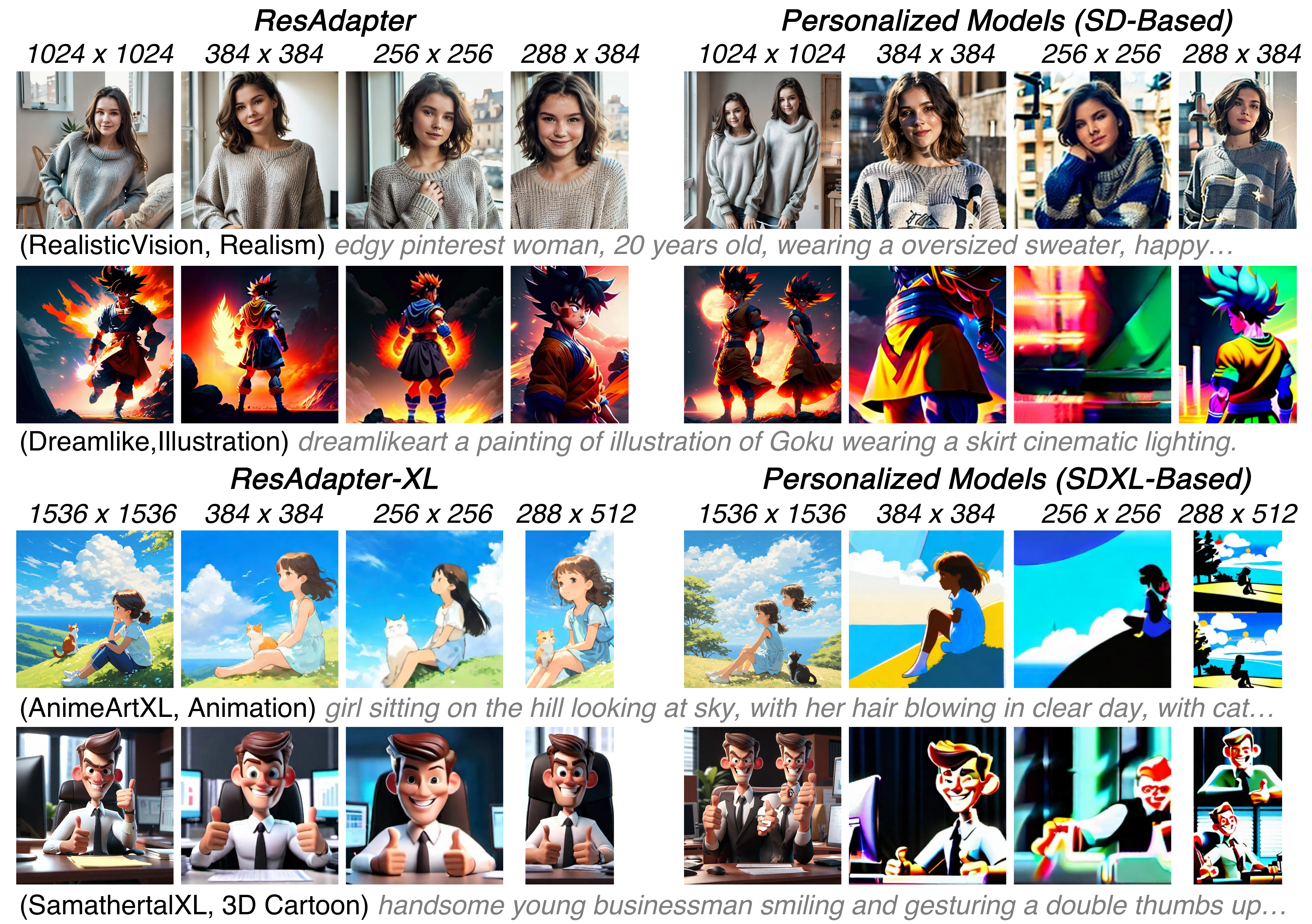}
  \caption{
  \textbf{Qualitative results}. 
  We compare the multi-resolution images generated by ResAdapter and the personalized models of arbitrary style domains. 
  \textbf{Left}: generation images from ResAdapter integrated into the personalized model.
  \textbf{Right}: generation images from the original personalized model. Some prompts are edited for clarity.
  }
\label{fig:fenggeduibi}
\vspace{-1.5em}    
\end{figure*}

\begin{table}[htb]
\centering
\resizebox{0.95\columnwidth}{!}{
    \begin{tabular}{@{}l|c|cc|c|c@{}}
    \toprule
    \textbf{Size} & \textbf{Method} & \textbf{FID($\downarrow$)} & \textbf{CLIP($\uparrow$)} & \textbf{Call} & \textbf{Latency/(s)} \\ \midrule
    \multirow{4}{*}{256x256} & $\text{SD}_{1.4}$ & 54.06 & 21.43 & 2 & 0.025\\
     & SDXL & 175.87 & 14.60 & 2 & 0.038 \\
     & $\text{ED}_{1.4}$ & 23.77 & 26.30 & 2 & 0.288 \\
     & $\textbf{Ours}_{1.4}$ & \bf{23.01} & \bf{26.98} & 2 & \bf{0.025}\\ \midrule
    \multirow{2}{*}{512x512} & $\text{SD}_{1.4}$ &20.50  &27.33  & 2 & 0.0714\\
     & $\textbf{Ours}_{1.4}$& 20.53& 27.32& 2 & 0.0714\\ \midrule
    \multirow{5}{*}{1024x1024} & SD & 47.01 & 25.70 & 2 & 0.1322\\
     & SDXL & 25.58 & \bf{28.06} & 2 & 0.1322\\
     & $\text{MD}_{1.4}$ & 37.70 & 26.96 & 162 & 2.50 \\
     & $\text{ED}_{1.4}$ & 27.76 & 26.07 & 33 & 1.16 \\
     & $\textbf{Ours}_{1.4}$ & \bf{26.89} & 27.26 & 2 & \bf{0.1322}\\ \bottomrule
    \end{tabular}
}
\caption{
    \textbf{Quantitative results} on LAION-COCO at the different resolutions. Call represents the number of inference iterations at each noise reduction step. For the latency, we measure the time of one step on an A100-80G.
}
\label{tab:laion}
\vspace{-1.5em}
\end{table}

\subsection{4.2\quad Main Results}
\label{mainresults}

\textbf{Comparison with Resolution-Free Generation Models.}
ResAdapter significantly improves the quality of multi-resolution images.
For \textit{quantitative results}, see Table~\ref{tab:laion}. 
The results show that ResAdapter outperforms MD and ED in terms of FID and CLIP Score.
About the latency time, ResAdapter without post-processing is, on average, $9 \times$ faster compared to ED. 
For \textit{qualitative results}, see more details in Appendix C.
We compare the image performance of MD and ED with our ResAdapter at the resolutions from 256 to 1024. 
The qualitative results demonstrate that ResAdapter generates multi-resolution images of better quality compared with MD and ED.
MD generates higher-resolution images with the poor framing and composition, and can not generate lower-resolution images than the training resolution. ED generates the images with more inference time.

\noindent \textbf{Comparison with Personalized Models.}
For \textit{quantitative results}, see Table~\ref{tab:gsb}. 
We evaluate the image quality by four criteria, which are the fidelity, the composition, the prompt alignment and the style domain consistency.
The quantitative results demonstrate that ResAdapter significantly outperforms the personalized model, particularly in lower-resolution images.
For \textit{qualitative results}, see Figure~\ref{fig:fenggeduibi}. To ensure the fairness of the experiments, we generate multi-resolution images using prompts from Civitai~\cite{Civitai}. These images are generated by ResAdapter and the personalized model. 
Lower-resolution images (e.g., $256 \times 256$, $384 \times 384$) generated by the personalized model are significantly lower in terms of the images of the fidelity, while higher-resolution images (e.g., $1536 \times 1536$) suffer from the poor framing and composition. 
After integrating our ResAdapter into the personalized model, the fidelity and the composition of generation images are significantly improved. 
ResAdapter enables the resolution extrapolation and interpolation of the personalized model.

\begin{figure*}[tb]

  \centering
  \includegraphics[width=0.95\linewidth]{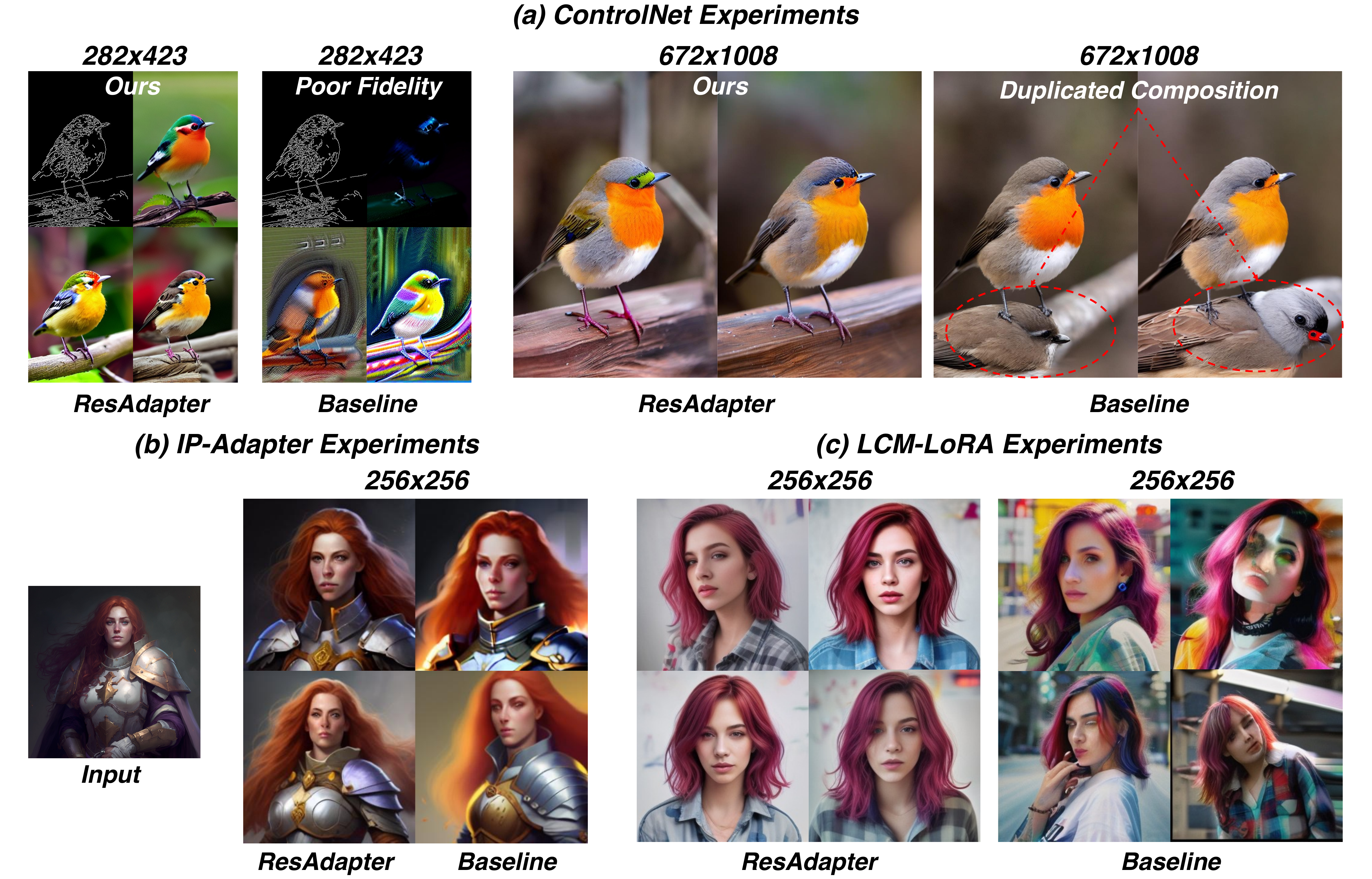}
  \caption{
  \textbf{Qualitative results} of extended experiments with ResAdapter. 
  Image-to-image tasks with \textbf{ConrtolNet}, the condition is canny images at different resolution.
  Image variation tasks with \textbf{IP-Adapter}, we resize the input image from 1024x1024 to 256x256.
 Accelerating text-to-image tasks with \textbf{LCM-LoRA}, we generate images in 4 steps.
  }
\label{fig:kuozhan1}
\vspace{-1.5em}
  
\end{figure*}

\begin{table}[htb]
\centering
\resizebox{0.95\columnwidth}{!}{
\begin{tabular}{@{}l|c|c|ccc|c@{}}
\toprule
\textbf{Domain}& \textbf{Size} & \textbf{Ratio} & \textbf{Good}& \textbf{Same} & \textbf{Bad} & \textbf{(G+S)/(B+S)}\\ \midrule
\multirow{4}{*}{Realism}& \multirow{2}{*}{256} & 4:3 &  3321& 1090& 589& 2.63\\
 &  & 3:4 & 3302& 1102& 596& 2.59\\ \cmidrule(l){2-7} 
 & \multirow{2}{*}{1024}& 16:9 & 2877 & 1563 & 560 & 2.09 \\
 &  & 9:16 & 2978 & 1531 & 491 & 2.23 \\ \midrule
\multirow{4}{*}{Illustration}& \multirow{2}{*}{256} & 4:3 &  3671& 1108& 221& 3.59\\
 &  & 3:4 & 3712& 982& 306 & 3.64\\ \cmidrule(l){2-7} 
 & \multirow{2}{*}{1024}& 16:9 & 2621 & 1762 & 617 & 1.84 \\
 &  & 9:16 & 2768 & 1598 & 634 & 1.95 \\ \midrule

\multirow{4}{*}{Animation} & \multirow{2}{*}{384} & 4:3 & 4479 & 402 & 119 & 9.36 \\
 &  & 3:4 & 4574 & 302 & 124 & 11.44 \\ \cmidrule(l){2-7} 
 & \multirow{2}{*}{1536} & 16:9 & 2978 & 1391 & 631 & 2.16 \\
 &  & 9:16 & 2809 & 1278 & 913 & 1.86 \\ \midrule
 
\multirow{4}{*}{Cartoon} & \multirow{2}{*}{384} & 4:3 & 4613 & 338 & 49 & 12.79 \\
 &  & 3:4 & 4684 & 249 & 67 & 15.61 \\ \cmidrule(l){2-7} 
 & \multirow{2}{*}{1536} & 16:9 & 2863 & 1566 & 571 & 2.07 \\
 &  & 9:16 & 3009 & 1071 & 920 & 2.05 \\ \bottomrule
\end{tabular}
}
\vspace{-0.5em}
\caption{
    \textbf{Quantitative results} by human for resolution-free generation of ResAdapter and personalized models.
}
\label{tab:gsb}
\vspace{-2em}
\end{table}

\begin{table}[htb]
\centering
\resizebox{0.95\columnwidth}{!}{
\begin{tabular}{@{}l|c|c|c|c@{}}
\toprule
\textbf{Method} & \textbf{Realism} & \textbf{Animation} & \textbf{Cartoon} & \textbf{Illustration} \\ \midrule
LoRA    & 35.67   & 51.28     & 47.81   & 45.18 \\
Diffit  & 31.34   & 47.53     & 41.24   & 38.72 \\
$\text{Norm}_{\text{Full}}$  & 31.34   & 47.53     & 41.24   & 38.72\\ \midrule
$\text{Ours}_{\text{ResENorm}}$    & 2.18    & 3.76      & 2.97    & 2.81 \\
$\text{Ours}_{\text{ResCLoRA}}$    & 2.29    & 3.69      & 3.02    & 2.84 \\
$\text{Ours}_{\text{Full}}$    & 2.32    & 3.94      & 3.08    & 2.89 \\ \bottomrule
\end{tabular}
}
\caption{
    \textbf{Quantitative results} about style consistency preservation. We train models on LAION-5B for the same epochs, and use FID to measure the gap from the distribution domain of the original persionalized model.
 }
\label{tab:style_consistency}
\vspace{-1.5em}
\end{table}

\noindent \textbf{Consistency with Personalized Models.}
We report the quantitative evaluation of ResAdapter's performance in maintaining style domain consistency when paired with personalized models, as summarized in Table~\ref{tab:style_consistency}. They undergo training for 10,000 iterations on the LAION-5B dataset, generating 5,000 images at training resolution within the personalized model framework. Unlike alternative approaches which exhibit significant divergence from the original style, ResAdapter preserves the integrity of the style domain. This evidence supports the conclusion that ResAdapter effectively upholds the stylistic characteristics of personalized diffusion models without degradation.

\begin{figure*}[tb]

  \centering
  \includegraphics[width=0.95\linewidth]{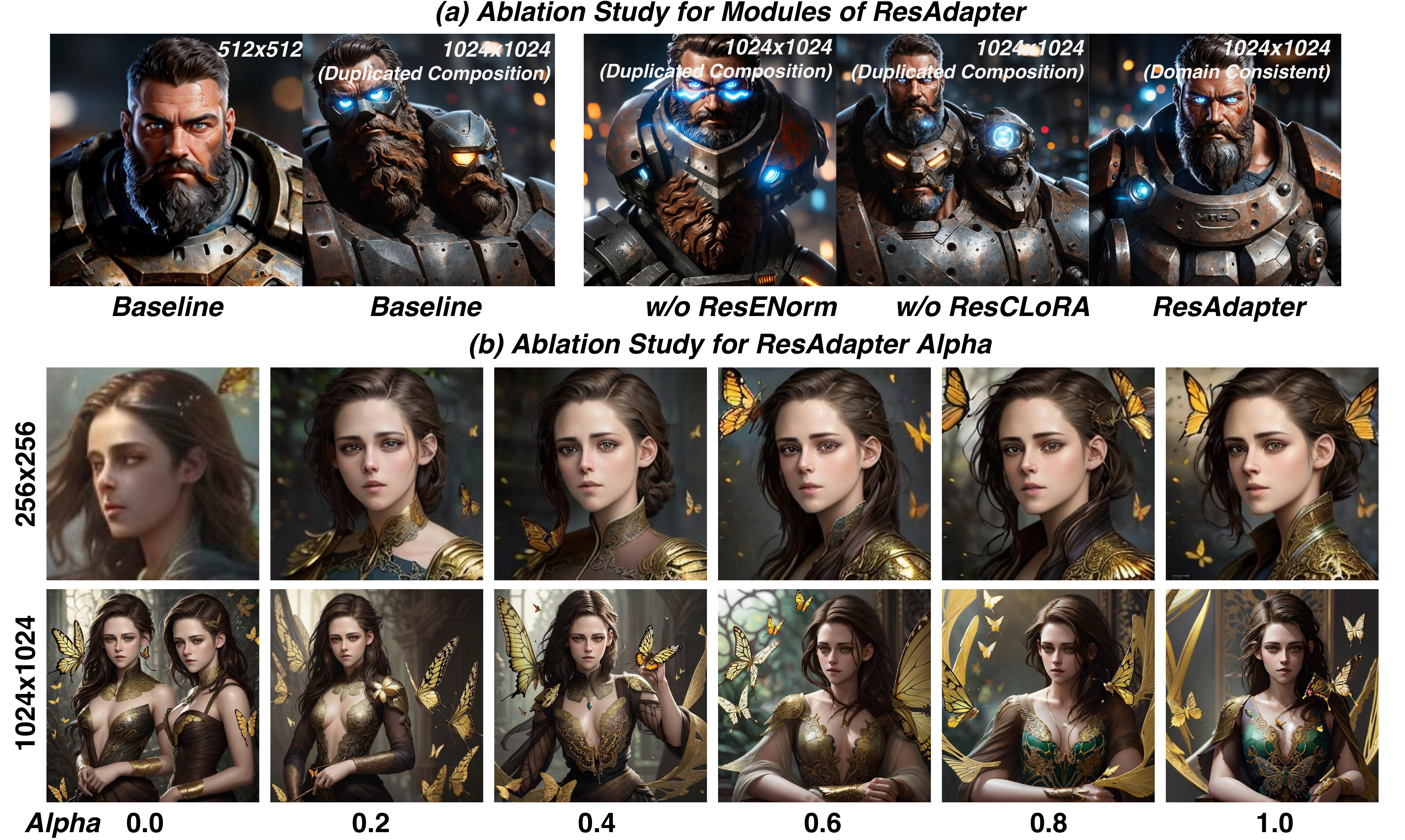}
  \caption{
  \textbf{Ablation studies.} 
  \textbf{Top}: We ablate on the modules of ResAdapter. Baseline represents Dreamshaper, which is a personalized diffusion model based on SD1.5. The third column represents only ResCLoRA integrated into the model. The fourth column represents only ResENorm integrated into the model. The fifth column represents both them integrated into the model.
  \textbf{Bottom:} We ablate on the alpha of ResAdapter $\alpha_r$ from 0 to 1 at lower and higher resolutions.
  }
  \label{fig:xiaorong}
  \vspace{-1.5em}
\end{figure*}

\subsection{4.3\quad Extended Results}
\label{extendres}

\textbf{ResAdapter with ControlNet.} 
ControlNet~\cite{zhang2023adding} is a conditional control module that can utilize conditional images to control the generation of layout-specific images for SD~\cite{rombach2022high}. As shown in Figure~\ref{fig:kuozhan1}, ControlNet 
generates low-quality images with poor fidelity and composition in the image-to-image task. 
While ResAdapter is compatible with ControlNet to enable the resolution extrapolation and interpolation, improving the quality of images.

\noindent \textbf{ResAdapter with IP-Adapter.}
IP-Adapter~\cite{ye2023ip} is a adapter for image generation with the image prompt.
As shown in Figure~\ref{fig:kuozhan1}, ResAdapter with IP-Adapter can generate high-quality images in the image variation task.

\noindent \textbf{ResAdapter with LCM-LoRA.}
LCM-LoRA~\cite{luo2023lcmlora} is a module for accelerated image generation capable of generating high-quality images in 4 steps. 
As shown in Figure~\ref{fig:kuozhan1}, ResAdapter is integrated into the personalized model and compatible with LCM-LoRA.
ResAdapter improves the fidelity of lower-resolution image while not degrades the quality of $512 \times 512$ images.

\noindent \textbf{ResAdapter with ElasticDiffusion.}
ResAdapter combined with other multi-resolution image generation models of post-processing can optimize the inference time.
Specifically, ED~\cite{haji2024elastic} requires to inference the $1024 \times 1024$ images multiple times, and overlaps them to get the $2048 \times 2048$ images with the post-process technology, which takes much inference time.
In order to optimize the inference time, we combine ResAdapter with ED to inference the $768 \times 768$ images same times and overlap them to get $2048 \times 2048$ images. ResAdapter with ED can generate $2048 \times 2048$ images while no degradation of image quality compared with ED. Our ResAdapter can speed up the inference time by 44\%, which can be found in Appendix C.
This demonstrates that ResAdapter can be flexibly applied to multiple scenarios.

\subsection{4.4\quad Ablation Studies}
\label{ablation}

For the modules of ResAdapter, we ablate on ResCLoRA and ResENorm, as shown in Figure~\ref{fig:xiaorong}-a.
Without ResCLoRA or ResENorm, the duplicated composition of generation images still exists, which demonstrate the importance of their simultaneous presence. Compared with baseline, ResAdapter can generate images without transforming the original style domain.
For alpha $\alpha_{r}$ for ResAdapter,  we make the ablation study on $\alpha_{r}$, as shown in Figure~\ref{fig:xiaorong}-b. We find the quality of generation images increases as $\alpha_{r}$ from 0 to 1.

\section{5\quad Conclusion}
\label{sec:conclusion}

In this paper, we present a plug-and-play domain-consitent ResAdapter for diffusion models of arbitrary style domain, which enables the resolution extrapolation and interpolation of generation images.
Our experiments demonstrate that after a low-cost training, ResAdapter with only 0.5M can be integrated into diffusion models to generate high-quality images of unrestricted resolutions and aspect without transforming the original style domain.
Our extended experiments also demonstrate ResAdapter is compatible with other modules (e.g., ControlNet, IP-Adapter and LCM-LoRA). 
In addition, ResAdapter can be combined with other multi-resolution image generation models (e.g., ElasticDiffusion) to optimize inference time for higher-resolution images.

%

\bibliography{aaai25}

\clearpage
\appendix
\begin{figure*}[htb]
  \centering
  \includegraphics[width=\linewidth]{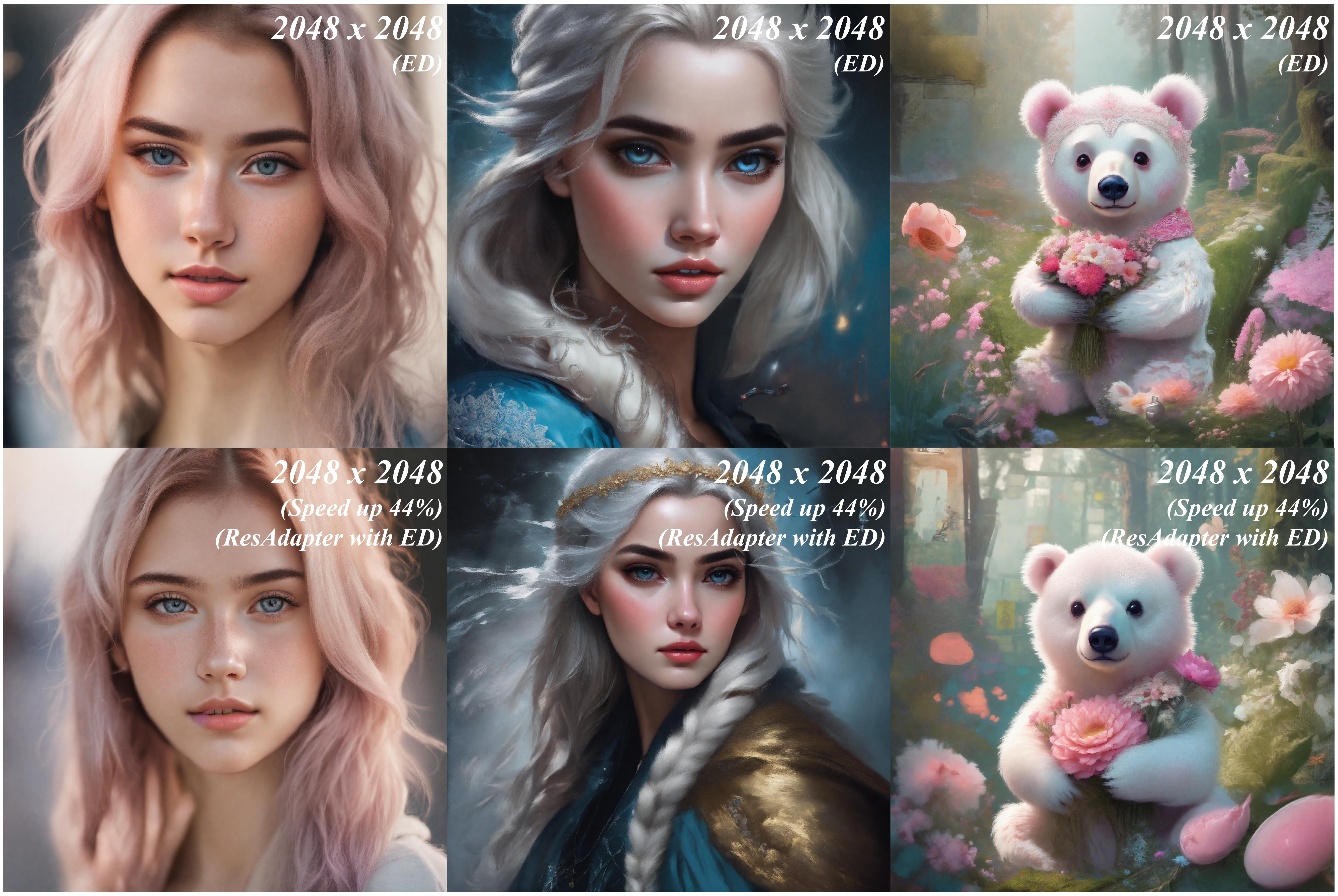}
  \caption{
  \textbf{Qualitative results} about ResAdapter with ElasticDiffusion (ED). 
  \textbf{Top:} DreamshaperXL (ED) generates $2048 \times 2048$ images by the post-process technology and uses $1024 \times 1024$ as the standard resolution.
  \textbf{Bottom:} DreamshaperXL (ResAdapter with ED), generates $2048 \times 2048$ images by the post-process technology and uses $768 \times 768 $ as the standard resolution.
  }
  \label{fig:train_free}

\vspace{-1em}
  
\end{figure*}

\section{A.\quad Mathematical Derivation of DMs}

The generation process of the diffusion model involves both forward diffusion and reverse denoise processes. 
Given a data sample $ x_0 \sim q_{data}(x) $, the diffusion model gradually injects small Gaussian noise into the data and generates samples through reverse denoise. 

\subsection{A.1\quad Forward Process}
Specifically, the \textit{forward diffusion process} of the diffusion model is controlled by a Markov chain as $ q(x_t|x_{t-1})=\mathcal{N}(x_t; \sqrt{1-\beta_t} x_{t-1}, \beta_t I) $, where $ \beta_t $ is a variance schedule between 0 and 1.
The data distribution $ q_{data}(x) $ can be transformed into a marginal distribution $q(x_t|x_0)$ by employing reparameterization tricks. Using the notation $\alpha_t := 1 - \beta_t$ and $\bar{\alpha}_t := \prod_{s=1}^{t} \alpha_s $, we have $q(x_t|x_0)=\mathcal{N}(x_t;\sqrt{\bar{\alpha}_t}x_0, (1-\bar{\alpha}_t)I)$.

\subsection{A.2\quad Reverse Process}
Considering the \textit{reverse process}, the diffusion model learns to progressively reduce small Gaussian noise as: $p_{\theta}(x_{t-1}|x_t)=\mathcal{N}(x_{t-1}; \mu_{\theta}(x_t,t), \sigma_t^2 I) $, where $\mu_{\theta}(x_t, t)=\frac{1}{\sqrt{\alpha_t}}(x_t-\frac{\beta_t}{\sqrt{1-\bar{\alpha}_t}}\epsilon_\theta(x_t,t)$.
The corresponding objective function is the variational lower bound of the negative log-likelihood. We get the equation as $\mathcal{L}(\theta) = \sum_t \mathcal{D}_{\text{KL}} (q(x_{t-1}|x_t,x_0)|p_\theta(x_{t-1}|x_t)) - p_\theta(x_0|x_1)$, where $ \mathcal{D}_{\text{KL}} $ denotes the KL divergence between the distribution $ p $ and $ q $. Furthermore, through the parameterization $\mu_\theta(x_t,t)$, the loss function can be simplified as:
\begin{align}
 L_{\text{simple}} = \mathbb{E}_{x_0,\epsilon,t}[||\epsilon - \epsilon_\theta(\sqrt{\bar{\alpha}_t}x_0 + \sqrt{1-\bar{\alpha}_t}\epsilon, t)||^2]
 \end{align}
The training objective in this formulation is minimizing the squared error between the gaussian noise and the estimated noise of the noise-added samples.

\section{B.\quad Experimental Details}
\subsection{B.1\quad Inference Details}
Our inference experiments encompass standard text-to-image (T2I) generation tasks, accelerated text-to-image generation tasks, and controllable image-to-image 
 (I2I) generation tasks. In our extended experiments, we have integrated tasks for generating images using accelerated LoRA~\cite{hu2021lora}, ControlNet~\cite{zhang2023adding}, and IP-Adapter~\cite{ye2023ip}. Additionally, we have also included tasks to generate images with other train-free resolution models.

Specifically, for the standard T2I generation task, the sampler utilized is DDIM~\cite{song2020denoise} with a guidance scale of 7.5 and an inference step count of 25. For the accelerated image generation task, the sampler employed is LCM~\cite{luo2023latent,song2020score}, with a guidance scale of 1.0 and an inference step count of 4. In the context of the controllable image-to-image generation task, the sampler for ControlNet is UniPC~\cite{zhao2024unipc}, with a guidance scale of 5.0, an inference step count of 16, and a denoising strength of 0.8. The injection strength for IP-Adapter is set at 0.85, with the remaining inference configurations being consistent with those of the standard T2I generation task.

For personalized diffusion models, we randomly selected prompts from CivitAI~\cite{Civitai} to generate images. For base models, we randomly chose prompts from LAION-COCO~\cite{schuhmann2021laion} to facilitate image generation. The definition of personalized models can be found in Appendix~B.2.

For tasks compatible with train-free resolution models, ResAdapter accelerates the generation of high-resolution images by reducing the resolution of image patches. ElasticDiffusion (ED)~\cite{haji2024elastic} generates images of $2048 \times 2048$ from $1024 \times 1024$ patches, while ResAdapter-ED produces high-resolution images from $768 \times 768$ patches.

Our primary evaluation focuses on the fidelity and style consistency of multi-resolution images. We evaluated the fidelity of the images using metrics such as FID~\cite{heusel2017gans}, and CLIP Score~\cite{hessel2021clipscore} to measure the extent to which ResAdapter can rectify bad cases across different resolutions. To evaluate the style consistency of images, we employ FID to gauge the degree of stylistic influence that ResAdapter exerts on personalized models.

\begin{table}[tb]
\centering
\resizebox{0.95\columnwidth}{!}{
\begin{tabular}{@{}l|l|l|l@{}}
\toprule
\textbf{Domain}       & \textbf{Name}            & \textbf{Type}  & \textbf{Train}      \\ \midrule
Stylistic    & Dreamshaper     & SD1.5 & Dreambooth \\
Stylistic    & Dreamshaper-XL  & SDXL  & Dreambooth \\
Illustration & Dreamlike       & SD1.5 & Dreambooth \\
Realism      & RealisticVision & SD1.5 & Dreambooth \\
Realism      & Juggernaut-XL   & SDXL  & Dreambooth \\
Cute         & Cuteyukimix     & SD1.5 & Dreambooth \\
Cartoon      & Samaritan-XL    & SDXL  & LoRA       \\
Animation    & Animeart-XL     & SDXL  & LoRA       \\ \bottomrule
\end{tabular}
}
\caption{Personalized diffusion models from CivitAI.}
\label{tab:personalized_diff}
\vspace{-1em}
\end{table}

\subsection{B.2\quad Personalized Diffusion Models}
\label{sec:pdm}

Personalized models refer to diffusion models that are specifically trained for a particular style domain using techniques such as  LoRA~\cite{hu2021lora} or Dreambooth~\cite{ruiz2023dreambooth}.
To validate the capabilities of ResAdapter, we selected eight personalized diffusion models spanning various styles, including realistic, anime, cartoon, 2.5D, and 3D styles. These models were sourced from Civit.ai, and detailed information can be found in Table~\ref{tab:personalized_diff}.

\section{C.\quad Compatible Qualitative Results}



The qualitative results for image generation using the compatible integration of ResAdapter and ElasticDiffusion are illustrated in Figure~\ref{fig:train_free}. This integration yields a 44\% speed enhancement over ED, while preserving the integrity of the ultra-high-resolution image generation performance.

\section{D.\quad Discussion about Our Work}



With the rapid advancement of diffusion models, we have observed a shift in the predominant architecture from UNet~\cite{ronne2015unet} to DiT~\cite{peebles2023scalable}. The primary motivation behind ResAdapter is to analyze the impact of convolutional and attention layers in UNet on image resolution. However, we have also conducted exploratory experiments on DiT, discovering that normalization layers can enhance the performance of high-resolution images, a conclusion that aligns with that of DifFit~\cite{xie2023difffit}. Nevertheless, a critical challenge arises as directly training normalization layers can lead to style conflicts. In the future, we aim to further explore how to learn low-resolution information on DiT and develop training strategies to prevent the model from learning style and semantic information.

\end{document}